\documentclass[]{opt2025} 
\usepackage{float}
\usepackage{amsmath}
\usepackage[
  backend=biber,
  style=authoryear,
  citestyle=authoryear,
  maxcitenames=1,      
  mincitenames=1,
  uniquelist=false     
]{biblatex}
\usepackage{wrapfig}
\usepackage{booktabs}
\newcommand{\nsharp}{\bar{\lambda}} 

\optauthor{%
\Name{Gunbir Singh Baveja} \Email{gbaveja@student.ubc.ca}\\
\addr University of British Columbia
\AND
\Name{Alex Lewandowski} \Email{lewandowski@ualberta.ca}\\
\addr University of Alberta
\AND
\Name{Mark Schmidt} \Email{schmidtm@cs.ubc.ca}\\
\addr University of British Columbia, Canada CIFAR AI Chair (Amii)
}

\title{A Unified Noise-Curvature View of Loss of Trainability}

\begin{document}
\maketitle

\vspace{-6mm}   
\begin{abstract}
Loss of trainability refers to a phenomenon in continual learning where parameter updates no longer make progress on the optimization objective, so accuracy stalls or degrades as the learning problem changes over time.
In this paper, we analyze loss of trainability through an optimization lens and find that the phenomenon is not reliably predicted by existing individual indicators such as Hessian rank, sharpness level, weight or gradient norms, gradient-to-parameter ratios, and unit-sign entropy.
Motivated by our analysis, we introduce two complementary indicators: a batch-size-aware gradient-noise bound and a curvature volatility-controlled bound.
We then combine these two indicators into a per-layer adaptive noise threshold on the effective step-size that anticipates trainability behavior.
Using this insight, we propose a step-size scheduler that keeps each layer’s effective parameter update below this bound, thereby avoiding loss of trainability.
We demonstrate that our scheduler can improve the accuracy maintained by previously proposed approaches, such as concatenated ReLU (CReLU), Wasserstein regularizer, and L2 weight decay.
Surprisingly, our scheduler produces adaptive step-size trajectories that, without tuning, mirror the manually engineered step-size decay schedules.\looseness=-1
\end{abstract}

\section{Introduction}
In continual learning, a model must sustain performance over a non-stationary data stream, meaning that optimization methods must remain effective as the model's parameters evolve. A central failure mode in continual learning is \emph{loss of trainability (LoT)}, in which parameter updates become increasingly ineffective, so training slows or stalls even when capacity and supervision are adequate \parencite{Lyle2023,Dohare2024}. This phenomenon is a major impediment to learning from new data, and is distinct from catastrophic forgetting, which is concerned with the retention of what was previously learned.
Empirically, LoT has been observed across architectures, datasets, and non-stationarities (e.g., random labels, permuted inputs, or incremental classes), suggesting it is a general property of gradient dynamics in non-stationary settings rather than an isolated peculiarity.\looseness=-1

Mechanistically, several indicators have been proposed as explanations for LoT: increasing weight magnitudes \parencite{Dohare2024}, decaying Hessian rank \parencite{Lewandowski2023}, reductions in gradient noise \parencite{Berariu2021}, dead units due to saturation \parencite{Dohare2024, Abbas2023}, unit linearization \parencite{Lyle2024Disentangling}, or unit sign entropy collapse \parencite{Lewandowski2024}. Yet in practice, these indicators are not reliable predictors of the phenomenon across optimizers and hyperparameters. A complete explanation that accounts for hyperparameters is particularly important because continual learning typically precludes \emph{lifetime tuning}: we cannot reset or retune step-sizes and regularizers for each task sequence \parencite{Ash2020,Mesbahi2025}. A robust indicator must therefore account for variability across hyperparameters, and it must explain the cases for which a single indicator fails to reliably predict LoT.\looseness=-1

In this work, we adopt a two-signal view of trainability. We propose a \emph{batch-size--aware} gradient-noise signal and a \emph{sharpness volatility} signal based on the normalized (Adam-adjusted) sharpness.
Together, these two signals yield a per-layer adaptive noise threshold on the effective step-size.
We use this insight to implement a simple per-layer scheduler for Adam that keeps each layer’s effective step below this bound.
This ensures updates are neither \emph{gradient-noise dominated} (where gradient randomness overwhelms the signal) nor \emph{curvature-noise dominated} (where sharpness volatility causes steps to alternate between descent and instability).

To evaluate the curvature volatility diagnosis and our scheduler as a mitigator, we consider a continual learning setting where \emph{lifetime tuning} of hyperparameters is infeasible.
We then consider continual learning algorithms that surface complementary failure routes: CReLU activations \parencite{Shang2016CReLU}, which ensure that every unit has a path for gradient flow, but do not prevent curvature collapse; Wasserstein regularizer \parencite{Lewandowski2023}, which curbs drift and stabilizes early tasks but degrades on longer horizons without step-size control; and standard L2 weight decay, the simplest scale regularizer for ReLU networks. We demonstrate that our scheduler improves the accuracy maintained by these approaches and, without tuning, yields adaptive step-size trajectories that mirror manually engineered decay schedules.

\vspace{-1mm}
\section{Related Work}\label{sec:related}
\vspace{-1mm}

\paragraph{Diagnosing loss of trainability}
A growing body of work characterizes LoT as a byproduct of stochastic gradient descent (SGD) that emerge when training on non-stationary data. Several studies emphasize geometric collapse: \textcite{Lewandowski2023} argue that reduced directions of curvature explain the onset of LoT, connecting Hessian rank degeneracy to a reduced capacity for learning. \textcite{Ziyin2024Symmetry} analyze symmetries in neural network objectives and show that SGD with L2 weight decay has a propensity to saddle points that results in sparse and low-rank solutions. Other accounts highlight instability caused by growing parameter norms \parencite{Dohare2024}, and more specifically growing sharpness \parencite{Foret2021SAM}. \textcite{Lyle2024Disentangling} found that layer normalization with growing parameter norms results in decaying effective step-sizes that reduce trainability. Beyond geometry and noise, unit activations have been implicated: saturation and linearization reduce the diversity of unit responses \parencite{Abbas2023,Lyle2024Disentangling}, which can be generalized to a collapse of unit-sign entropy \parencite{Lewandowski2024}. More broadly, multiple different potential mechanisms can contribute to LoT, depending on the specific configuration of the learning algorithm. As we will show, existing individual mechanisms provide an incomplete explanation for LoT.\looseness=-1

\paragraph{Step-size as noise injection and plasticity.}
The step-size of an optimization algorithm plays a central role in plasticity: it acts as controllable noise \parencite{Mandt2017, SmithLe2017}, and regulates the balance between escaping sharp minima and settling into stable solutions. While classical noise-scale theory prescribes step-sizes based on gradient signal-to-noise ratios \parencite{Schaul2013}, we find this insufficient in non-stationary settings where curvature volatility acts as a distinct, dominant failure mode.
Our results refine this view in continual learning: larger step-sizes are not uniformly beneficial. When gradients become noise-dominated or curvature becomes fragile, \emph{decaying} the step-size restores trainability. For instance, under L2 and Wasserstein regularizer, we often reduce the base step-size before briefly re-warming, whereas CReLU tolerates mild warm-ups only after volatility drops (Sec.~\ref{sec:bound+controller}). This aligns with widely used decaying schedules \parencite{Goyal2017LargeBatch,Loshchilov2017SGDR,Huang2017Snapshot,Vaswani2017Attention,Devlin2019BERT,Chen2017RethinkingAtrous}.\looseness=-1

\begin{figure}
\centering
\begin{minipage}{.33\linewidth}\centering
\includegraphics[width=1.8in]{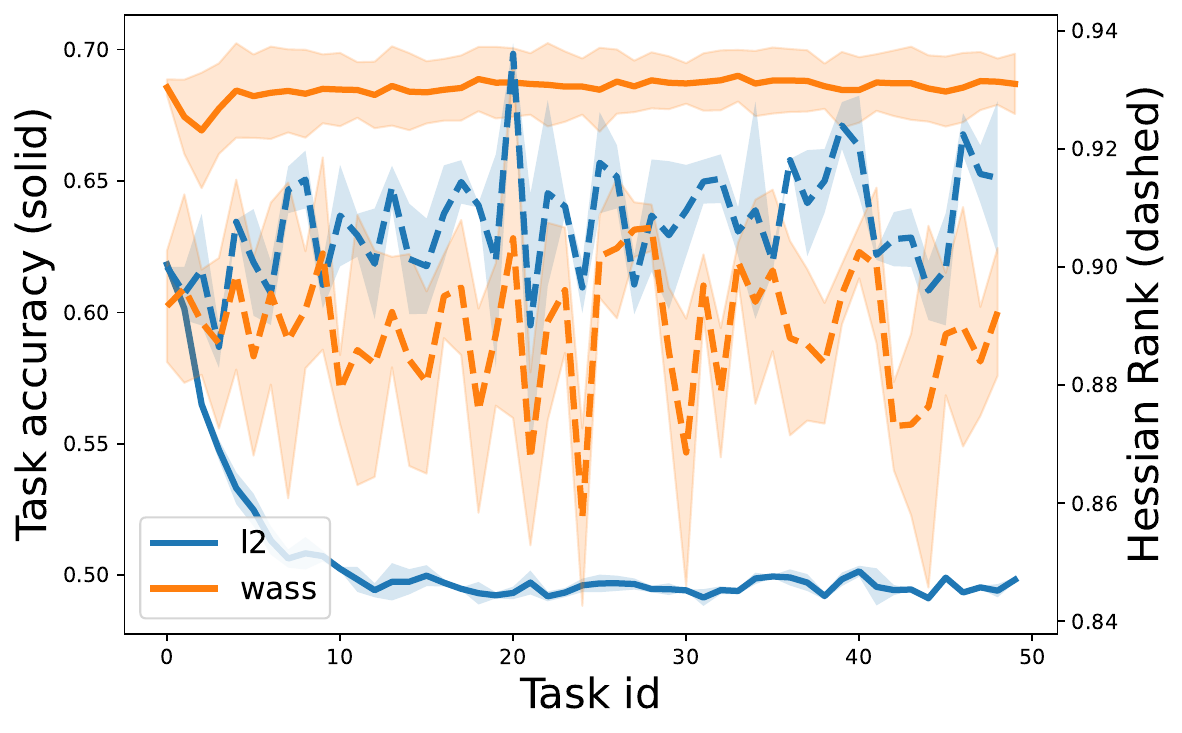}
\small (a) Hessian rank
\end{minipage}\hfill
\begin{minipage}{.33\linewidth}\centering
\includegraphics[width=1.8in]{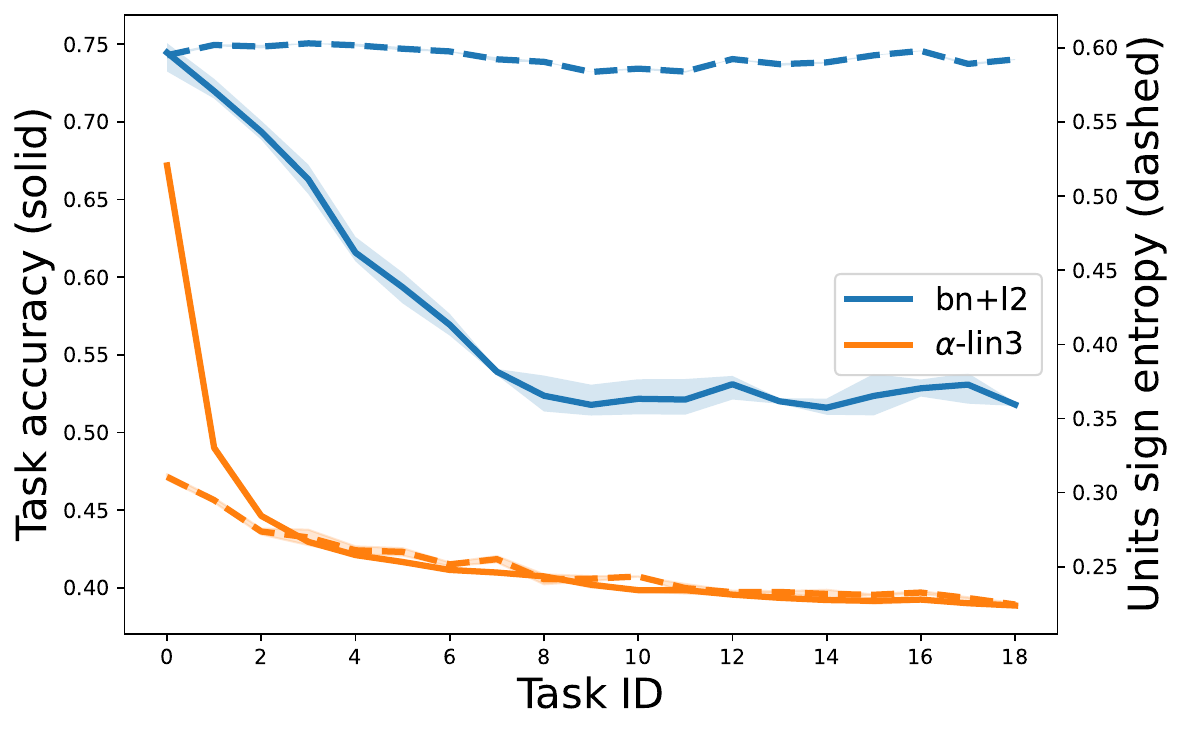}
\small (b) Unit-sign entropy
\end{minipage}\hfill
\begin{minipage}{.32\linewidth}\centering
\includegraphics[width=1.8in]{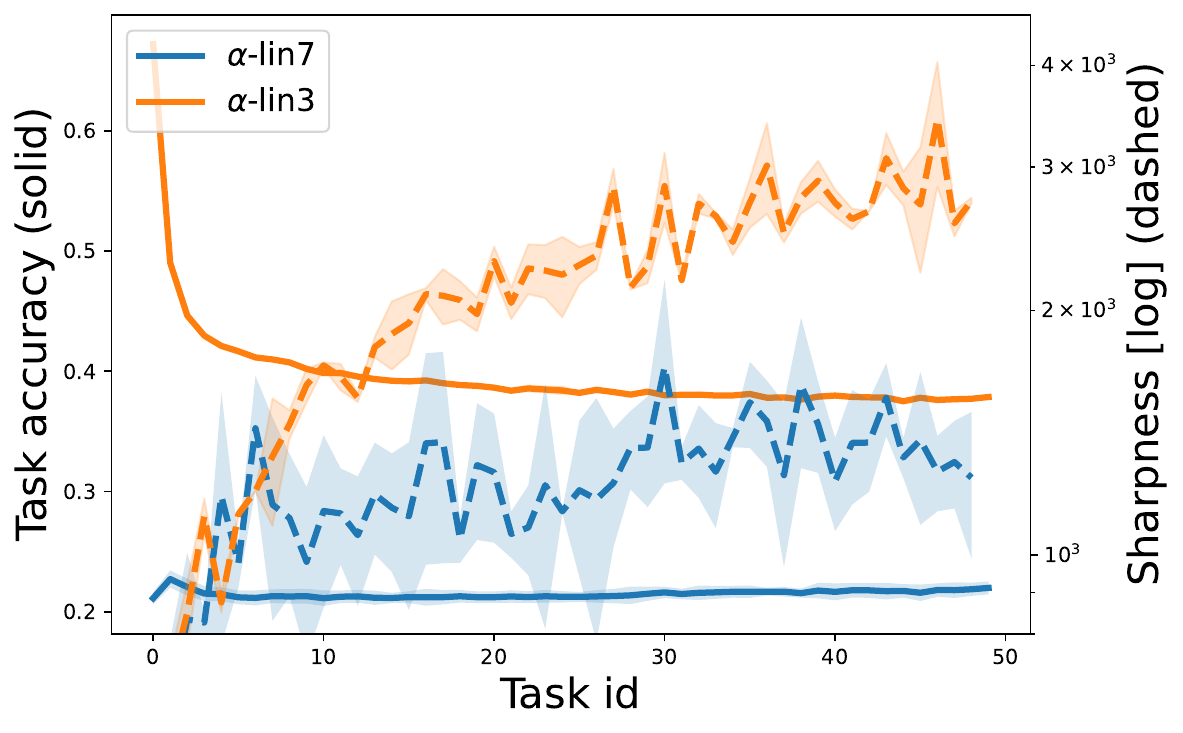}
\small (c) Sharpness
\end{minipage}\hfill
\begin{minipage}{.33\linewidth}\centering
\includegraphics[width=1.8in]{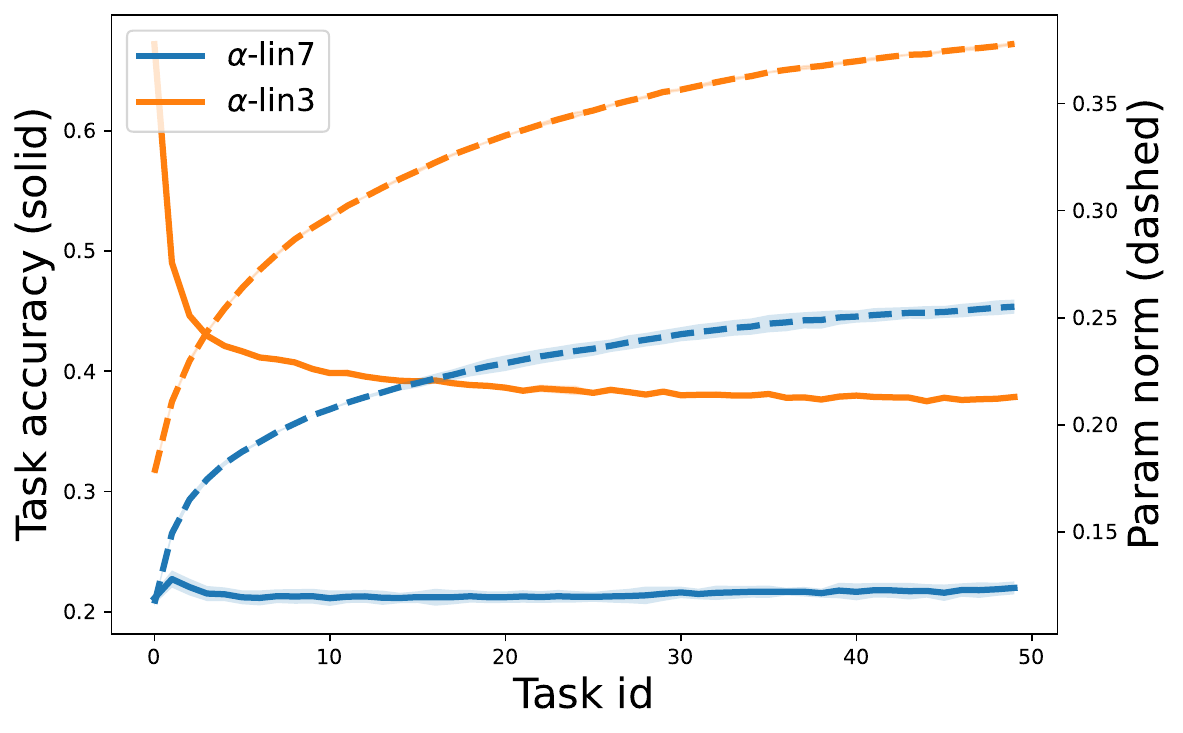}
\small (d) Weight norm
\end{minipage}
\begin{minipage}{.33\linewidth}\centering
\includegraphics[width=1.8in]{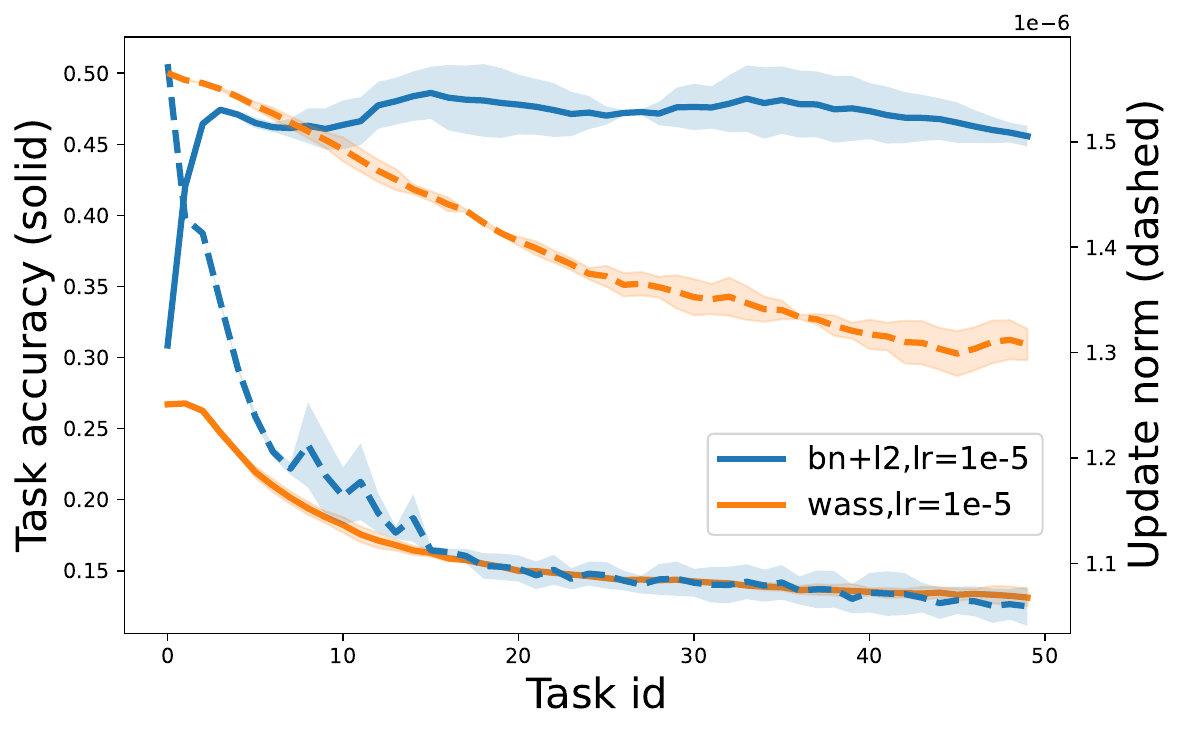}
\small (e) Grad norm
\end{minipage}
\begin{minipage}{.32\linewidth}\centering
\includegraphics[width=1.8in]{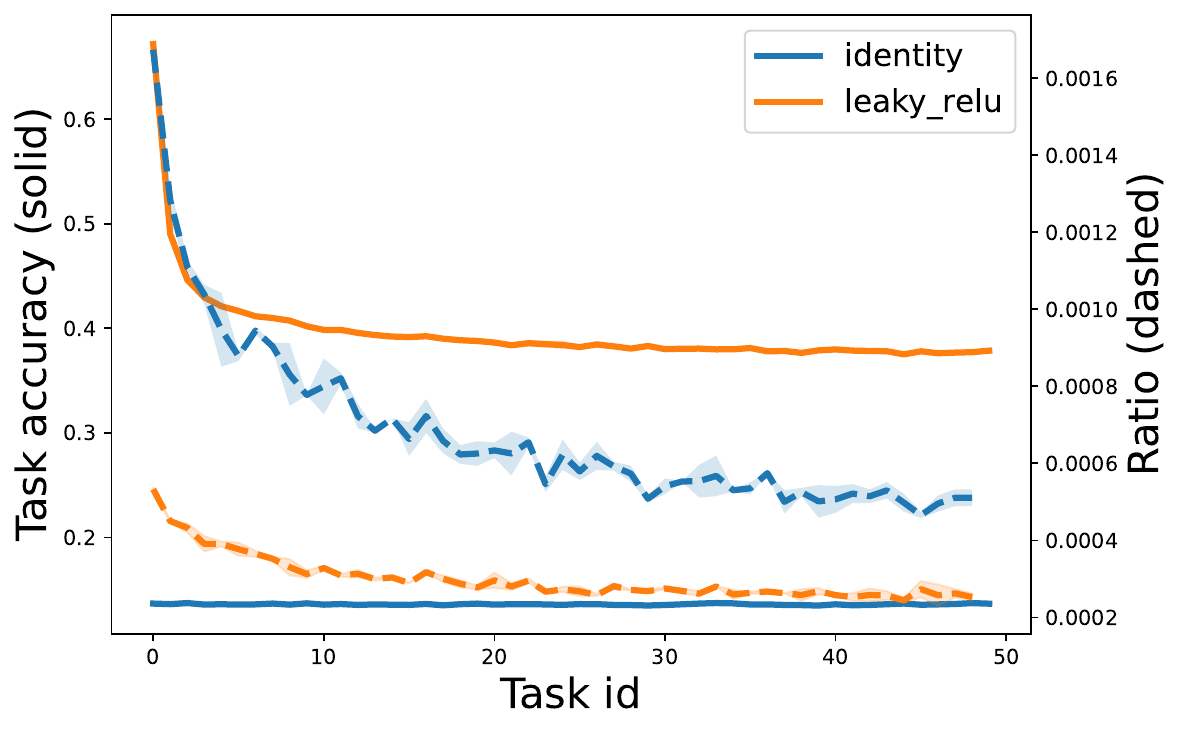}
\small (f) Grad/Param ratio
\end{minipage}
\caption{\small \textbf{Single indicators are incomplete explanations for loss of trainability.} 
In each panel, we compare two configurations (blue vs.\ orange). 
If an indicator were reliable, a collapse in the metric (dashed) should consistently predict a collapse in accuracy (solid).
However, we observe contradictions; for example, in \textbf{(a)}, the L2 model maintains high Hessian rank yet suffers catastrophic accuracy loss, whereas the Wasserstein model maintains accuracy despite similar rank behavior.
}
\label{fig:failures}
\vspace{-1.5em}
\end{figure}

\vspace{-1mm}
\section{No Complete Single Explanation}\label{sec:fail}
\vspace{-1mm}
Here we present experimental findings in which no existing individual indicator provides a complete explanation of LoT across architectures or hyperparameters (see Appendix \ref{app:setup} for details).
We see that in Fig.~\ref{fig:failures}:
\emph{(a) Hessian rank:} High rank is often associated with trainability, yet under L2 weight decay, the model sustains a high rank but suffers a total collapse in accuracy ($\approx 0.1$). Conversely, the Wasserstein regularizer maintains high task accuracy ($\approx 0.65$) despite exhibiting a similarly stable rank profile.
\emph{(b) Unit-sign entropy:} While low entropy (feature collapse) can signal LoT, BatchNorm with weight decay maintains stable entropy yet loses trainability. In contrast, LeakyReLU ($\rho{=}0.3$) shows a drop in entropy but preserves better trainability.
\emph{(c) Sharpness:} Both LeakyReLU ($\rho{=}0.7$) and LeakyReLU ($\rho{=}0.3$) exhibit rising sharpness, yet only the former suffers catastrophic trainability loss.
\emph{(d) Weight Norm:} Comparing the same pair of networks, they show similar growth in weight norm, but exhibit differing trainability behavior.
\emph{(e) Gradient Norm:} Gradient decay appears in both the failing configuration (BatchNorm with weight decay) and the successful configuration (Wasserstein regularizer), yet the latter, with faster decay, remains trainable.
\emph{(f) Grad/Param Ratio:} This ratio falls for both deep linear and LeakyReLU networks, but only the latter loses trainability.

\vspace{-1mm}
\section{Two Signals that Unify \& Predict LoT}\label{sec:signals}
\vspace{-1mm}
Not all drops in task accuracy indicate the same underlying failure. We observe two distinct routes to trainability loss: (i) phases in which updates are \emph{noise-dominated} so progress stalls despite adequate curvature, and (ii) phases in which curvature collapses \parencite{Ziyin2024Symmetry, Lewandowski2023} or becomes \emph{noisy}, so nominally similar steps flip between descent and instability.

\paragraph{Critical step-size from gradient noise}
We analyze the conditions for expected descent along the update direction (see derivation in Appendix~\ref{app:theo-grad_noise_proof}).
While Adam determines the update direction via preconditioning, the step-size magnitude determines the regret guarantee \parencite{kingma14_adam}. 
In general, the optimal choice of step-size depends on both the curvature of the optimization landscape and the gradient noise \parencite{Schaul2013}.
Assume the loss $J$ is locally $L$-smooth along the update direction.
Let the minibatch gradient be $\widehat g_t = g_t + \xi_t$, where $g_t=\mathbb{E}[\widehat g_t]$ is the true gradient, $\sigma^2_{t,\mathrm{ps}}$ a per-sample variance proxy, and $B$ the batch size.
A quadratic expansion of $J$ yields the sufficient condition for expected descent:
\[
\alpha_t \;<\; \frac{2}{L}\cdot\frac{\|g_t\|^2}{\|g_t\|^2+\sigma^2_{t,\mathrm{ps}}/B}
\;\le\; \frac{2}{L}\frac{B\|g_t\|^2}{\sigma^2_{t,\mathrm{ps}}}.
\]
This recovers the optimal step-size form derived by \textcite{Schaul2013}, adapted here to the batch-size--aware setting.
Defining the noise-critical step-size
\(\alpha_g^\star(t):=B\|g_t\|^2/\sigma^2_{t,\mathrm{ps}}\),
we obtain the compact criterion
$\alpha_t \;\le\; 2\,\alpha_g^\star(t)/L$.
Empirically, \(\alpha^{\star}_{g}\) hovers near \(1\) early (see \(\alpha^{\star}_{g}\) in Fig.~\ref{fig:alpha-safe}) and becomes more noisy late in tasks. As the effective step-size \(\alpha_t\) drifts upward, the fraction of steps where \(\alpha_t\!>\!\alpha^{\star}_{g}(t)\) grows. This noise-dominated phase is evident in L2 where accuracy deteriorates although the curvature signal remains consistent. (center in Fig.~\ref{fig:alpha-safe}).\looseness=-1

\begin{figure}[t]
\centering
\includegraphics[width=0.85\linewidth]{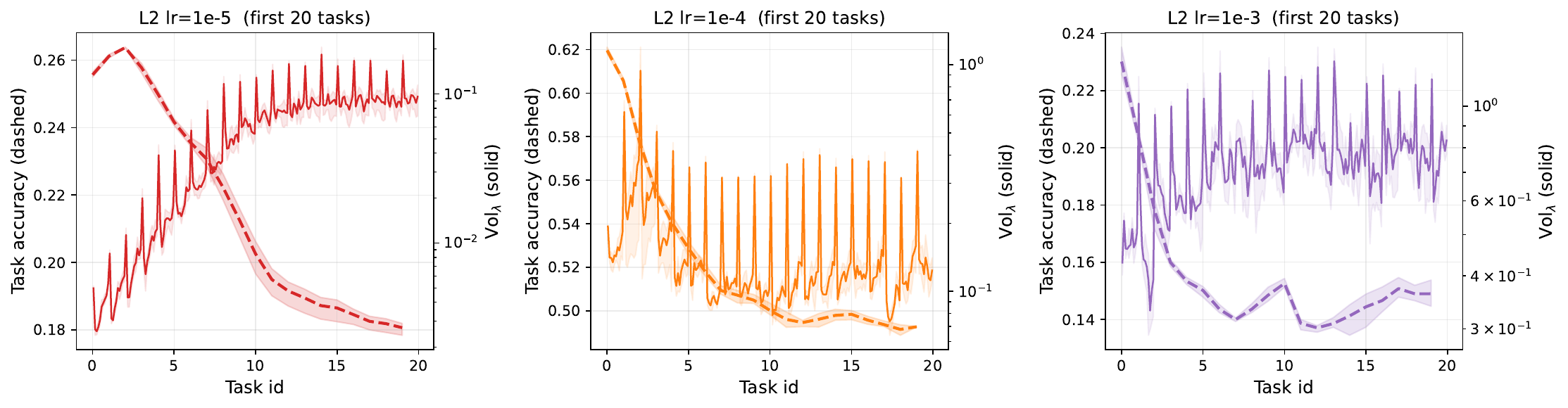}
\vspace{-1.5em}
\caption{\small \textbf{Sharpness volatility as an indicator.} Under L2 regularization, the rise in curvature volatility (solid) consistently precedes and mirrors the collapse in task accuracy (dashed).}
\vspace{-1em}
\label{fig:early-warning}
\end{figure}

\paragraph{Critical step-size from sharpness volatility}
We typically lack access to the true Hessian at every step, and must instead rely on stochastic estimates.
To account for Adam's preconditioning, we track the \emph{normalized} sharpness \(\bar{\lambda}_t\) (the curvature projected along the preconditioned update direction, see Appendix~\ref{app:theory-norm}).
We estimate the raw top eigenvalue $\lambda_t$ efficiently at every step via power iteration using a single Hessian-vector product, incurring negligible overhead (see Appendix~\ref{app:impl-metrics}).
Since curvature fluctuates during training, we estimate its temporal signal-to-noise ratio.
We use an exponential moving average mean \(\mu_t\) and variance \(\sigma_t^2\) of \(\bar{\lambda}_t\) to define the dimensionless \emph{curvature volatility}:\looseness=-1

\begin{equation}\label{eq:voli}
\mu_t:=\frac{(1 - \nu) \mu_{t-1} + \nu\bar{\lambda}_t}{1-\nu^t},\quad
\sigma_t^2:=(\bar\lambda_t-\mu_t)^2,\quad
\mathrm{Vol}_{\bar{\lambda}}(t):=\frac{\sigma_t^2}{\mu_t+\varepsilon}\,.
\end{equation}
High volatility indicates that the local curvature is noisy or changing rapidly, destabilizing large steps.
In Appendix~\ref{app:volatility}, we use a quadratic expansion to show that ensuring stability with high probability requires the step-size to scale inversely with this volatility. Capturing the constant scaling factors in $\kappa$, we obtain the bound: 
\(\;
\alpha_t \;\le\; \alpha^{\star}_{\mathrm{vol}}(t) \;:=\; \left({\kappa\,\mathrm{Vol}_{\bar{\lambda}}(t)}\right)^{-1}\,.
\)

This bound tightens as sharpness variance exceeds average sharpness, which we refer to as fragile (or unstable) sharpness.

\emph{Interpretation.} The bound shows that larger $\mu_t$ and smaller $\sigma_t^2$ make curvature along the preconditioned direction more \emph{predictable}, permitting larger safe steps. 
In continual learning, volatility can arise when the optimizer moves through non-stationary landscapes with shifting geometry or slips into sharp regions where curvature estimates fail. This pattern has been observed empirically when weight-norm growth or feature collapse drives the optimizer into regions of the parameter-space where local geometry varies drastically across mini-batches.
Consequently, when $\alpha^{\star}_{\mathrm{vol}}(t)$ \emph{decays} ($\sigma_t^2\gg\mu_t$), the curvature signal weakens; persistent exceedance $\alpha_t>\alpha^{\star}_{\mathrm{vol}}(t)$ indicates curvature-noise dominated updates and precedes LoT (Fig.~\ref{fig:early-warning}).

\paragraph{Effective step-size drift}
Even with L2 weight decay, the layerwise effective step-size \(\alpha_t\) drifts upward across tasks (Fig.~\ref{fig:alpha-safe}). 
As \(\alpha_t\) approaches or crosses thresholds \(\alpha^{\star}_{g}(t)\) and \(\alpha^{\star}_{\mathrm{vol}}(t)\), an increasing share of updates becomes gradient noise- or curvature-noise dominated. In principle, LoT can be mitigated by reducing \(\alpha_t\) or, equivalently, by increasing the batch size (\(B\)).

\begin{figure}[t]
\centering
\includegraphics[width=0.85\linewidth]{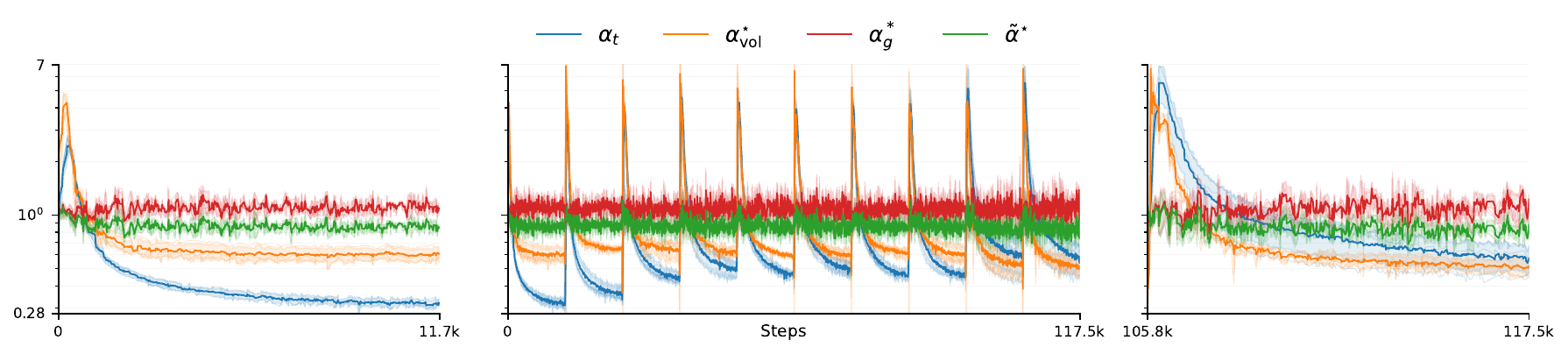}
\vspace{-1.3em}
\caption{\small \textbf{First layer step-size dynamics when training with L2 weight decay ($\lambda=10^{-3}$).} The effective step-size $\alpha_t$ (blue) exhibits upward drift across tasks due to weight-norm growth, which Adam's preconditioning converts into an increased step-size.}
\vspace{-1em}
\label{fig:alpha-safe}
\end{figure}

\section{Adaptive Noise Threshold and Per-Layer Scheduler}\label{sec:bound+controller}
\begin{figure}[h]
\centering
\includegraphics[width=0.9\linewidth,height=1.1in]{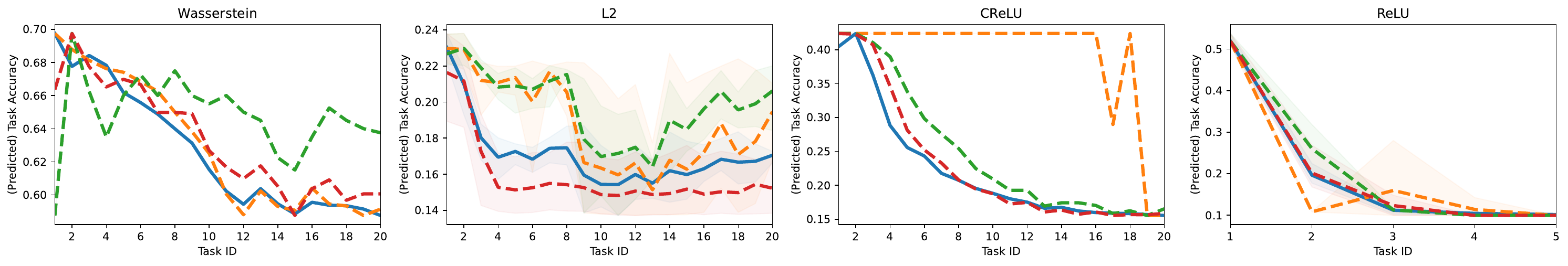}
\includegraphics[width=0.9\linewidth]{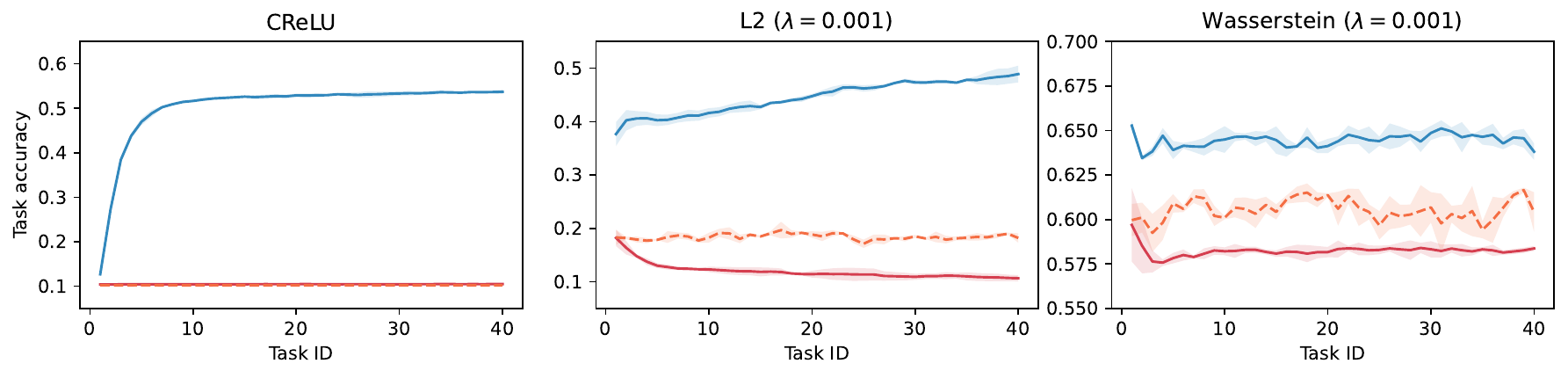}
\vspace{-1.5em}
\caption{\small \textbf{Diagnosis and Mitigation.} \textbf{Prediction (top):} The combined bound $\tilde{\alpha}^*$ (red) most accurately anticipates the onset of accuracy drops (blue) compared to individual metrics ($\alpha^*_{\mathrm{Vol}}$, green; $\alpha^*_g$, orange). \textbf{Performance (bottom):} The per-layer scheduler (blue) consistently restores trainability across all baselines, significantly outperforming vanilla (red) training and task resets (orange).\looseness=-1}
\vspace{-1em}
\label{fig:predict_controller}
\end{figure}

While classical noise-scale theory provides an upper bound on step-size based on gradient variance \parencite{Schaul2013}, we find it insufficient for non-stationary continual learning.
As shown in our ablation (Fig.~\ref{fig:ablation_study}), trainability is only consistently restored when satisfying a joint constraint.
We therefore combine the two critical effective step-sizes using a curvature volatility-aware noise proxy ($\tilde{\sigma}^2$). This tightens the bound precisely when curvature becomes fragile by inflating the effective noise.
We define this per-layer ($\ell$) noise proxy and the corresponding critical effective step-size as:\looseness=-1
\vspace{-0.5em}
\begin{align}
\tilde\sigma_{t}^{2(\ell)}
&:=\;
\sigma_{t,\mathrm{ps}}^{2(\ell)}
\;+\;
\beta\,\big\|\widehat g_t^{(\ell)}\big\|^2\,
\mathrm{Vol}_{\bar{\lambda}}^{(\ell)}(t),
\;\; \beta\in[0,1]\footnotemark;
\qquad
\widetilde{\alpha}_{t}^{\star(\ell)}
:=\;
\frac{B\,\big\|\widehat g_t^{(\ell)}\big\|^{2}}{\tilde\sigma_{t}^{2(\ell)}}\,.
\label{eq:alpha-tilde-layer}
\end{align}
\footnotetext{where $\beta$ can be tuned to control the inflation, which we set at $1$.\vspace{-1em}}
\vspace{-1em}
\paragraph{Simple adaptive per-layer scheduler}
From Fig.~\ref{fig:predict_controller} (top), we see that this combined bound predicts trainability behavior most consistently, where the predicted accuracy serves as a proxy for the frequency of step-size violations (see details in Appendix~\ref{app:impl-expl}). The scheduler tracks $\widetilde{\alpha}_t^{\star(\ell)}$ and defines a per-layer safe bound
$\alpha_t^{\mathrm{safe}(\ell)} := (1-\epsilon)\,\widetilde{\alpha}_t^{\star(\ell)}$.
Every $K$ steps, Adam’s per-layer effective step-size is adjusted: if $\alpha_t^{(\ell)}$ exceeds $\alpha_t^{\mathrm{safe}(\ell)}$ the layer’s base LR is cooled, and if it lies conservatively below this bound early in training, it is warmed. Refer to Appendix~\ref{app:impl} (Alg.~\ref{alg:scheduler}) for implementation details and Appendix~\ref{app:exp-depth} for an empirical comparison of the per-layer vs. full-network controller performance.\looseness=-1

\section{Experiments}\label{sec:experiments}
To evaluate the curvature volatility diagnosis and our scheduler as a mitigator, 
we investigate LoT on a sequence of 40 non-stationary tasks using a random-label MNIST.
Each task involves memorizing randomized target labels to isolate optimization dynamics from semantic transfer. 
We use a two-layer MLP (width 256) trained with Adam, where we use hyperparameters tuned for a single task (\emph{no lifetime tuning}).
More details on the setup in Appendix~\ref{app:setup}.


Figure~\ref{fig:predict_controller} (bottom) compares vanilla, reset-at-task, and our per-layer scheduler across methods.
With CReLU, our scheduler steadily improves on each task, whereas vanilla and reset stay near random accuracy.
We attribute this initial failure of the vanilla baseline to the absence of regularization constraints (present in L2 and Wasserstein), which otherwise implicitly moderate the effective step-size and prevent immediate entry into a noise-dominated regime (see Appendix~\ref{app:exp-depth}).
Under L2, our scheduler continues improving, while vanilla decays to random accuracy as the effective step-size gradually increases.
Lastly, Wasserstein regularization exhibits mild decay in trainability, yet our controller maintains a higher and more stable average task accuracy.

\begin{wrapfigure}[10]{r}{0.3\textwidth}
  \centering
  \includegraphics[width=0.85\linewidth]{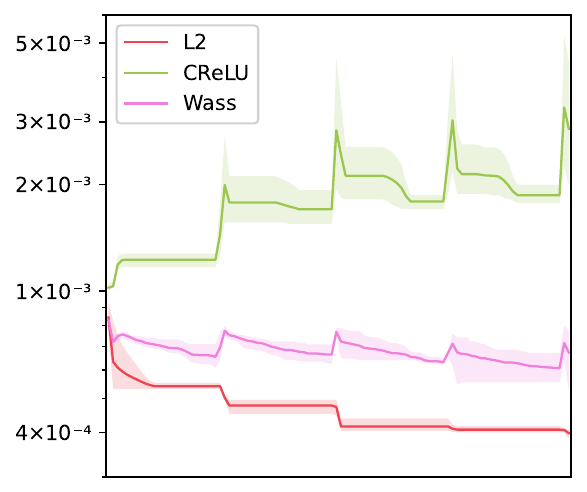}  
  \vspace{-1em}
  \caption{\small Scheduled step-sizes on the first 4 tasks.}
  \label{fig:lr-behavior}
  \vspace{-\baselineskip} 
\end{wrapfigure}
Figure~\ref{fig:lr-behavior} shows the step-size trajectories produced by our scheduler. 
The step-size decreases as training progresses within a task, mirroring standard schedules that are know to yield improved performance.
Moreover, the step-size schedule is adaptive to the effect of non-stationarity in combination with the learning algorithm.
For L2, the step-size cools sharply at the start and settles to a low plateau far below the initial $10^{-3}$; for Wasserstein, a brief warm-up is followed by per-task cool-downs. For CReLU, the step-size warms moderately, then plateaus with within-task cool-downs, matching the drop in sharpness volatility and the rise in accuracy.
These step-size trajectories are significant because they are effective at improving performance, are achieved without tuning, and validate our analysis of the loss of trainability phenomenon. 

\section{Acknowledgements}
The work was partially supported by the Canada CIFAR AI Chair Program and NSERC Discovery Grant RGPIN-2022-036669

\printbibliography

\appendix
\section{Experiments}\label{app:exp}
\subsection{Varying task-length}\label{app:exp-length}
\begin{figure}[H]
\centering
\includegraphics[width=0.9\linewidth]{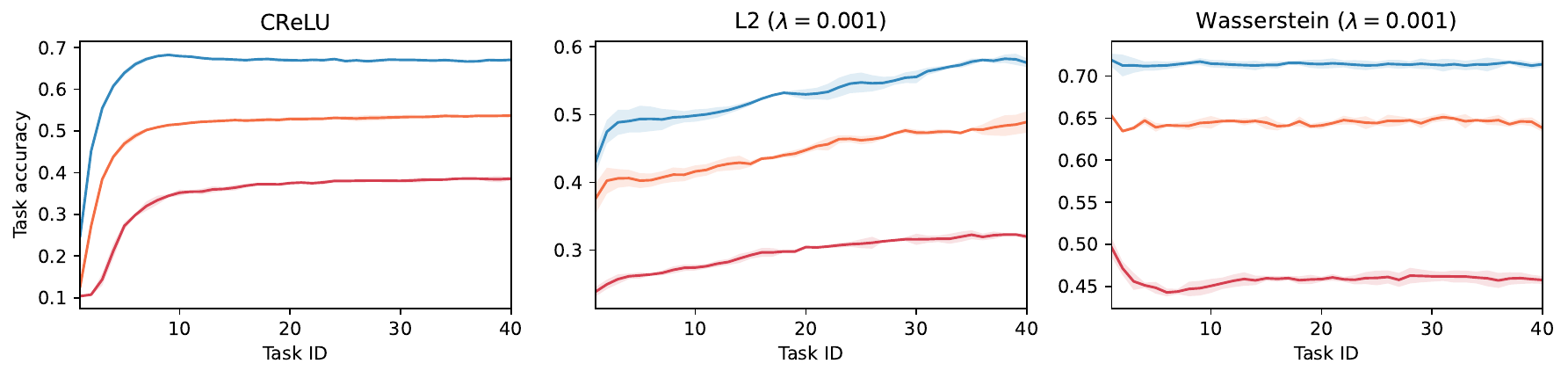}
\includegraphics[width=0.9\linewidth]{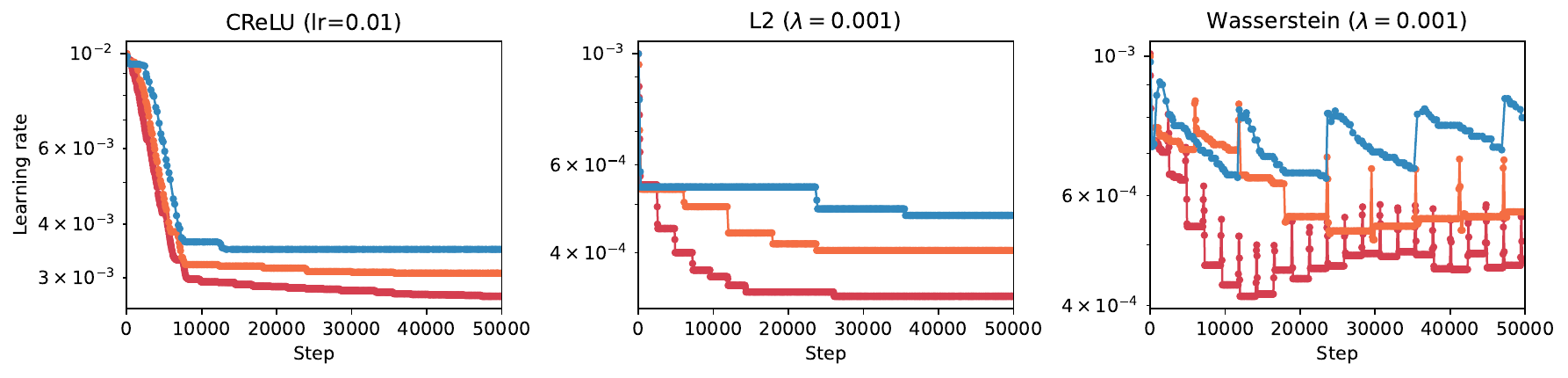}
\caption{\small \textbf{Task accuracy (top) and learning rate (bottom) behavior across methods, ablating over task-lengths.} Red: $100$ epochs/task. Orange: $250$ epochs/task. Blue: $500$ epochs/task. The scheduler successfully adapts to longer horizons, maintaining higher terminal learning rates and leveraging the increased budget to improve accuracy.}
\label{fig:task_accs_length_abl}
\end{figure}

In this section, we analyze the sensitivity of our proposed scheduler to the duration of tasks. Figure~\ref{fig:task_accs_length_abl} illustrates the evolution of average task accuracy and the adaptive learning rate trajectories for CReLU, L2 weight decay, and Wasserstein regularizer across three distinct task lengths: 100, 250, and 500 epochs per task.

We observe that increasing the task horizon consistently leads to improved asymptotic performance across all methods. This behavior is notable because standard implementations of these baselines often degrade on longer horizons; without adaptive step-size control, the accumulation of gradient noise and curvature volatility typically accelerates the onset of loss of trainability as the number of update steps increases. The monotonic improvement in accuracy (top row) suggests that our scheduler successfully decouples trainability preservation from task duration, allowing the model to effectively leverage the additional optimization budget.

The learning rate trajectories (bottom row) elucidate the mechanism behind this robustness. We observe that the adaptive schedules are qualitatively similar across task lengths but scale in magnitude. Specifically, for longer task durations (e.g., 500 epochs, shown in blue), the learning rate anneals less aggressively and stabilizes at a higher terminal value compared to shorter horizons (e.g., 100 epochs, shown in red). This indicates that the scheduler effectively identifies that the ``safe" effective step-size region remains accessible for longer periods when the optimization is not rushed, thereby maintaining a higher learning rate to maximize convergence without crossing the adaptive noise threshold.

\subsection{Ablation study}\label{app:exp-albation}
\begin{figure}[H]
\centering
\includegraphics[width=1.0\linewidth]{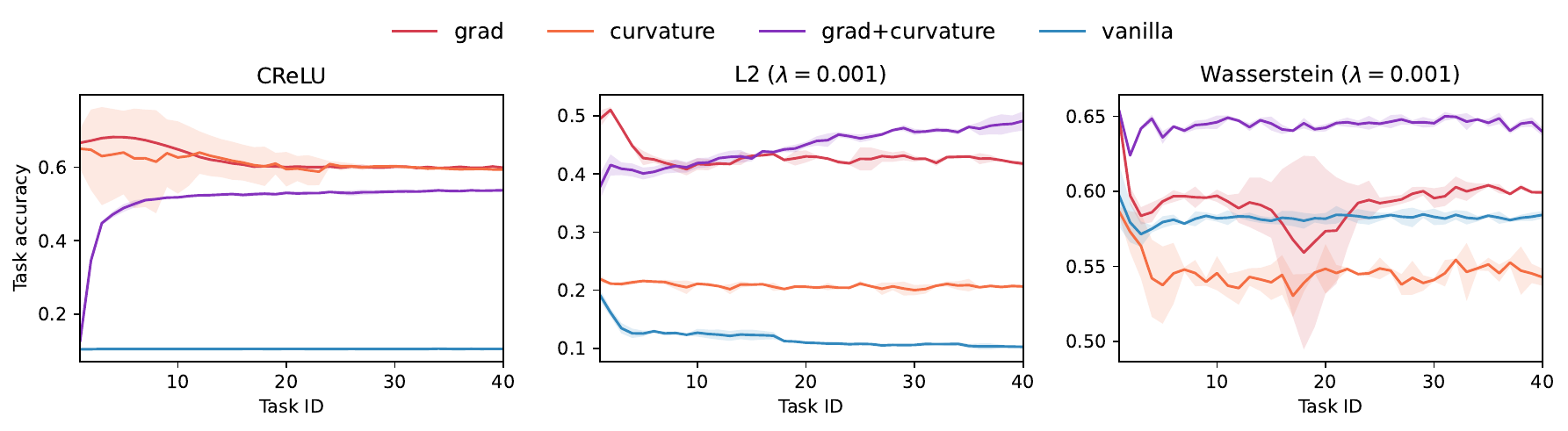}
\vspace{-2em}
\caption{\small \textbf{Impact of removing different signals from combined bound.} The combined gradient and curvature bound consistently yields superior and more stable trainability across CReLU, L2, and Wasserstein regularizer, highlighting the complementary roles of both signals.}
\label{fig:ablation_study}
\end{figure}

For CReLU, both grad-only (using $\alpha^*_g$) and curvature-only (using $\alpha^*_{\text{vol}}$) signals initially lead to a temporary loss of trainability before eventually stabilizing at a higher accuracy than vanilla or the combined method. This might suggest that relying on a single signal can lead to an overestimation of the safe step-size during initial tasks, causing transient instability.

\begin{figure}[H]
\centering
\includegraphics[width=0.9\linewidth]{fig/l2_fc1.pdf}
\includegraphics[width=0.9\linewidth]{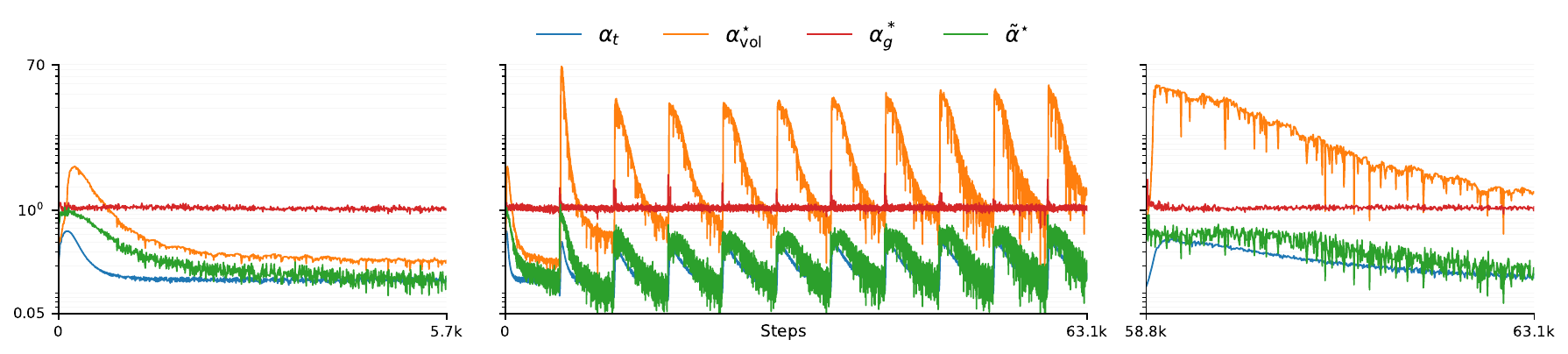}
\includegraphics[width=0.9\linewidth]{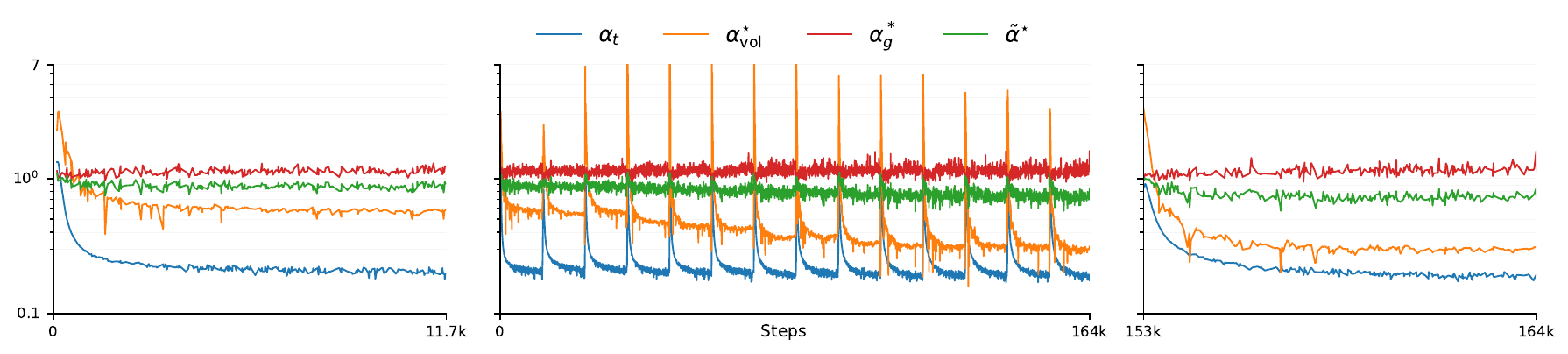}
\caption{\small \textbf{Effective step-size vs.\ sharpness-aware safe bound.}
We plot the first layer's effective step-size \(\alpha_{t,l_1}\) with \(\alpha^{\star}_{g,l_1}\), \(\tilde\alpha^{\star}_{l_1}\), and \(\alpha^{\star}_{\mathrm{vol},l_1}\) for L2 regularizer (top), CReLU (middle), and Wasserstein regularer (bottom) ($\lambda=10^{-3}$). Task \(1\) (left), Tasks \(1\!-\!10\) (middle), and Task \(10\) (right).}
\label{fig:alpha-safe-all}
\end{figure}

With L2 regularization, the grad-only-signal method incurs loss of trainability at the start, but is able to maintain a higher accuracy compared to vanilla. In contrast, the curvature-only signal shows a slight improvement over vanilla but does not fully  maximize performance. 
This behavior is elucidated by examining the critical step-size trajectories in Figure~\ref{fig:alpha-safe-all} (top row, L2 regularizer). The $\alpha^{\star}_{g}$ (red) bound for L2 often hovers at relatively high values, indicating that if only gradient noise is considered, the scheduler would perceive a safer regime than is actually present, leading to insufficient learning rate decay. The $\alpha^{\star}_{\mathrm{vol}}$ (orange) bound, while more responsive, might preclude optimal performance. The combined bound successfully navigates this by dynamically weighing both factors, leading to a significant and sustained increase in accuracy.

For Wasserstein regularizer, the ablation results are particularly insightful. The grad-only signal performs poorly,  the curvature-only signal (orange curve) yields performance worse than even the vanilla baseline after the initial tasks. This suggests that the $\alpha^{\star}_{\mathrm{vol}}$ signal, when used in isolation for Wasserstein, may lead to excessive and premature learning rate decay, unduly limiting the model's capacity to learn. As observed in Figure~\ref{fig:alpha-safe-all} (bottom row, Wasserstein regularizer), $\alpha^{\star}_{\mathrm{vol}}$ is indeed much more unstable and generally lower than $\alpha^{\star}_{g}$. Surprisingly, we see that the combined threshold is also lower than the gradient-only threshold, yet the learning rate, when controlled to lower than the combined bound leads to better performance.

The combined-bound method demonstrates consistent effectiveness in preserving trainability and achieving a high and stable average task accuracy across the above regularization methods. 

\subsection{Per-layer vs Full-network}\label{app:exp-depth}
\begin{figure}[H]
\centering
\includegraphics[width=0.9\linewidth]{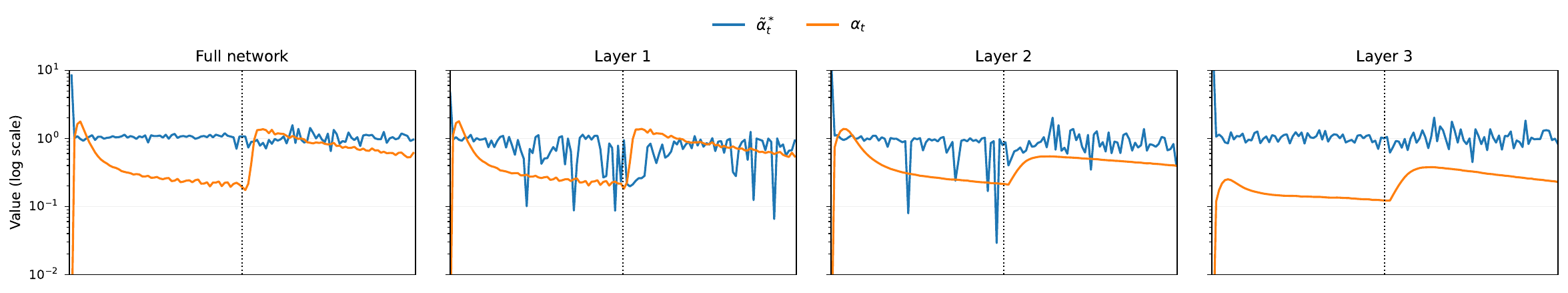}
\includegraphics[width=0.9\linewidth]{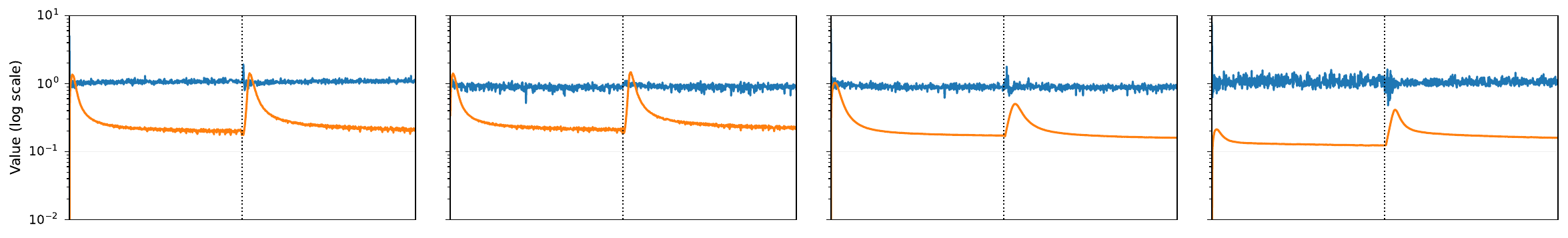}
\includegraphics[width=0.9\linewidth]{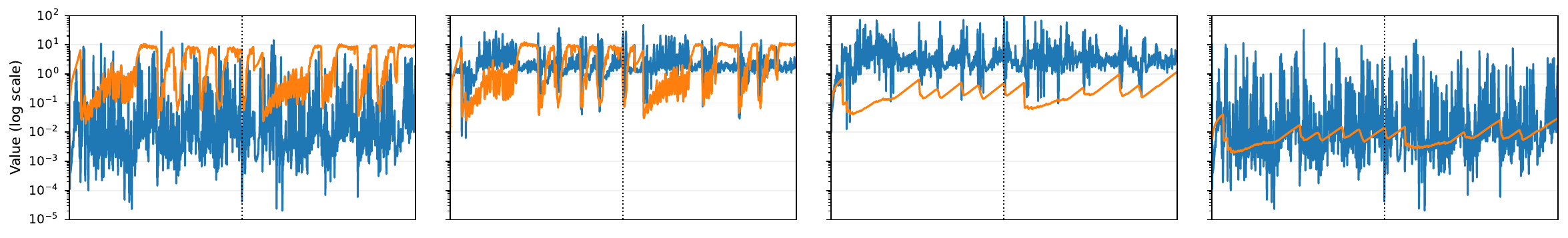}
\includegraphics[width=0.9\linewidth]{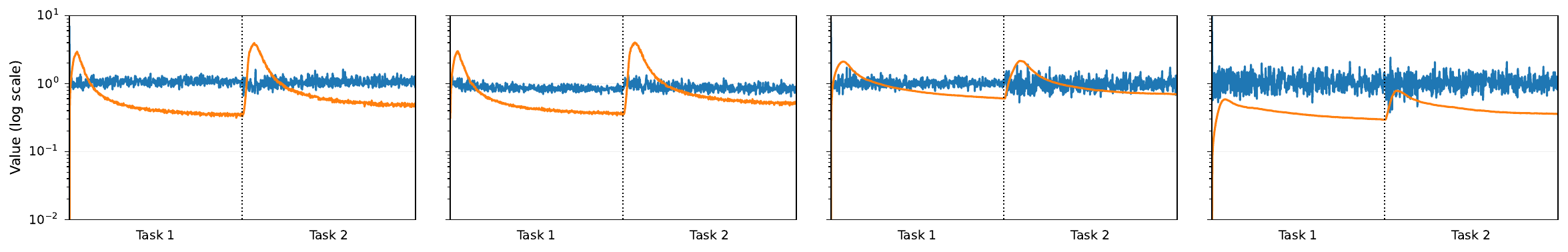}
\caption{\small \textbf{Layer behavior.} From top to bottom : ReLU, Wasserstein regularizer, CReLU, L2 regularizer.
The effective step-size $\alpha_t$ (orange) and critical bound $\tilde{\alpha}^*_t$ (blue) for the full network and individual layers. The global signal effectively mirrors Layer 1, often masking the stability of deeper layers.}
\label{fig:layer_behavior}
\end{figure}

In this section, we investigate whether loss of trainability is a global phenomenon or if it is localized to specific layers. Figure~\ref{fig:layer_behavior} presents a comparison between global (full-network) tracking and per-layer tracking of the effective step-size $\alpha_t$ against the critical stability bound $\tilde{\alpha}^*_t$.

\paragraph{Heterogeneity of layer dynamics (Figure~\ref{fig:layer_behavior}).}
The first column displays the metrics calculated over the full network parameters. We observe that the global signal (leftmost column) is heavily correlated with the dynamics of the first layer (second column). Specifically, when the full network suggests that $\alpha_t$ exceeds the critical bound $\tilde{\alpha}^*_t$, it is almost exclusively driven by the instability in Layer 1. We can also see (Fig.~\ref{fig:layer_behavior}) that for ReLU, at the start of task 2, the effective step size in Layer 1 is much higher above the threshold than is noted using the full network optimizer and sharpness statistics. In contrast, Layer 2 and Layer 3 often remain well within the safe regime (effective step-size below the critical threshold).
Intuitively, the first layer is directly exposed to the raw input stream and thus bears the immediate brunt of the covariate shift at task boundaries. Deeper layers operate on latent representations which may be more stable, or at least exhibit different sensitivity scales.

This analysis also sheds light on the early failure of the vanilla CReLU baseline noted in Section~\ref{sec:experiments}.
In the third row of Figure~\ref{fig:layer_behavior}, we observe that the critical stability bound ($\tilde{\alpha}^*_t$, blue) for CReLU is highly volatile and frequently drops below the effective step-size ($\alpha_t$, orange), especially in {Layer 1}.
In the absence of regularization (like L2 or Wasserstein) to implicitly constrain the step-size or smooth the curvature volatility, the vanilla optimizer operates in a violation regime ($\alpha_t > \tilde{\alpha}^*_t$) immediately.

\paragraph{Per-layer Calculation.}
To capture this heterogeneity, we compute the effective step-size $\alpha_{t}^{(\ell)}$ for a specific layer $\ell$ by averaging the element-wise Adam step multiplier across all parameters within that layer's top-level module. Formally, for a parameter $p$ at step $t$, the effective step is $ \eta{(1-\beta_1^t)^{-1}} / (\sqrt{\widehat{v}_t} + \epsilon) $. We aggregate this scalar value for all weights and biases within a layer to derive $\alpha_{t}^{(\ell)}$. This allows us to detect when a specific layer enters a curvature-noise-dominated regime, even if the global aggregate statistics suggest stability (or vice-versa).

\begin{figure}[H]
\centering
\includegraphics[width=0.9\linewidth]{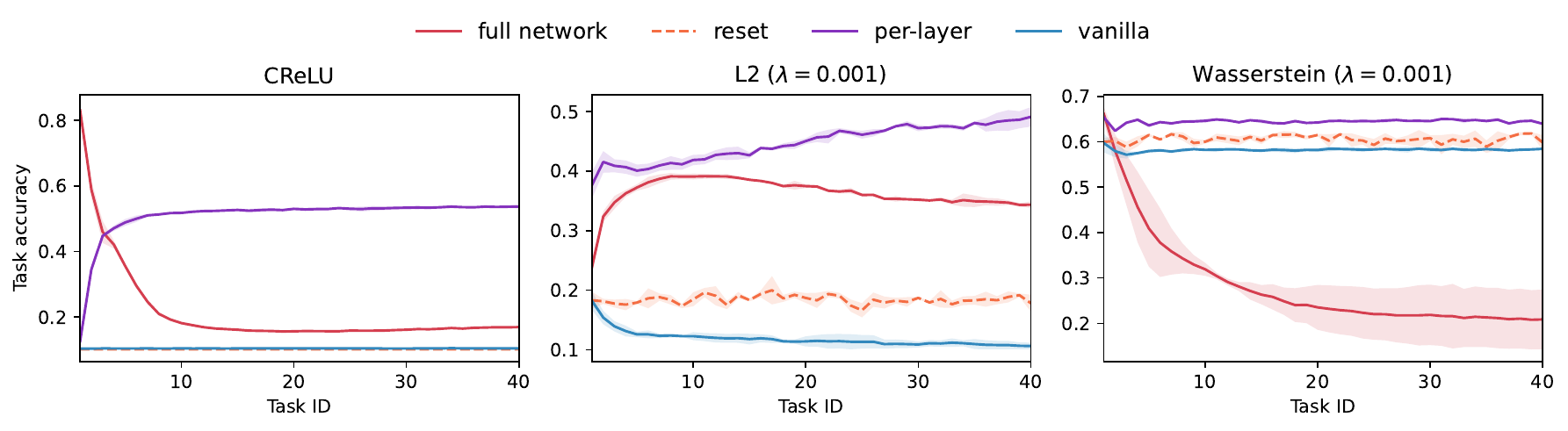}
\caption{\small \textbf{Per-layer vs.\ Full-network controller performance.} Task accuracy comparing a controller driven by full-network statistics (red) versus per-layer statistics (purple). The per-layer approach significantly outperforms the global approach.}
\label{fig:ly_v_pl}
\end{figure}

\paragraph{Per-layer vs. full-network control (Figure~\ref{fig:ly_v_pl}).}
The consequence of this heterogeneity is shown in the performance comparison. A global controller, which modulates the learning rate based on the full-network signal, is forced to penalize the entire model based on the worst-case behavior (typically Layer 1). This unnecessarily constrains deeper layers that are statistically healthy, reducing their capacity to learn.
By contrast, our per-layer controller (purple curve) acts independently: it cools the learning rate only for the specific layers violating their stability bounds (typically Layer 1 during high-volatility phases) while allowing stable layers (Layers 2 and 3) to maintain higher learning rates. This results in significantly higher task accuracy compared to the full-network controller (red curve), confirming that there is no single step-size that is optimal for all layers simultaneously.

\section{Theoretical details}\label{app:theory}
\subsection{Linking existing explanations}\label{app:theo-link}
In positively homogeneous networks, growth in L2/spectral weight norms increases network sensitivity and spectral norms (and typically loss sharpness), so both the top eigenvalue $\lambda_t$ and the \emph{normalized (Adam-adjusted)} sharpness $\nsharp_t$ (and its variability) rise. This elevates the volatility $\mathrm{Vol}_{\nsharp}(t)$ defined in Eq.~\eqref{eq:voli}, which \emph{shrinks} the curvature–volatility critical step-size
\[
\alpha_{\mathrm{vol}}^\star(t)\;=\;\frac{1}{\kappa\,\mathrm{Vol}_{\nsharp}(t)}\,,
\]
making updates brittle \parencite{Keskar2016,Santurkar2018,YoshidaMiyato2017,Foret2021}. 
Apparent ``decaying gradient norm" or a falling grad/param ratio are naturally interpreted through the \emph{batch-size–aware} critical step-size $\alpha_g^\star(t)$ together with the effective step-size $\alpha_t$: if $\|\widehat g_t\|$ shrinks while the per-sample variance stays high (or $\alpha_t$ drifts upward), a larger share of updates satisfies $\alpha_t>\alpha_g^\star(t)$, entering the noise-dominated regime without any necessary change in rank. Crucially, because LoT is generally assessed as a reduction in average task accuracy, perceived loss of trainability might occur without a measured effect in the ``indicators" if Adam functions in a regime of noise-dominated steps even when the curvature remains healthy (see $\alpha_t$ at the start of tasks in Fig.~\ref{fig:alpha-safe}).

Conversely, high Hessian rank can coexist with LoT when either \emph{noise-dominated} crossings $\alpha_t>\alpha_g^\star(t)$ or \emph{curvature-noise-dominated} crossings $\alpha_t>\alpha_{\mathrm{vol}}^\star(t)$ proliferate; low-rank regimes can remain trainable when both criteria are comfortably satisfied—explaining Fig.~\ref{fig:failures}. 
We refer the reader to Appendix~\ref{app:theo-hr} for a discussion of when Hessian rank decay aligns with these signals.
\vspace{1em}
\begin{figure}[H]
\centering
\includegraphics[width=0.9\linewidth]{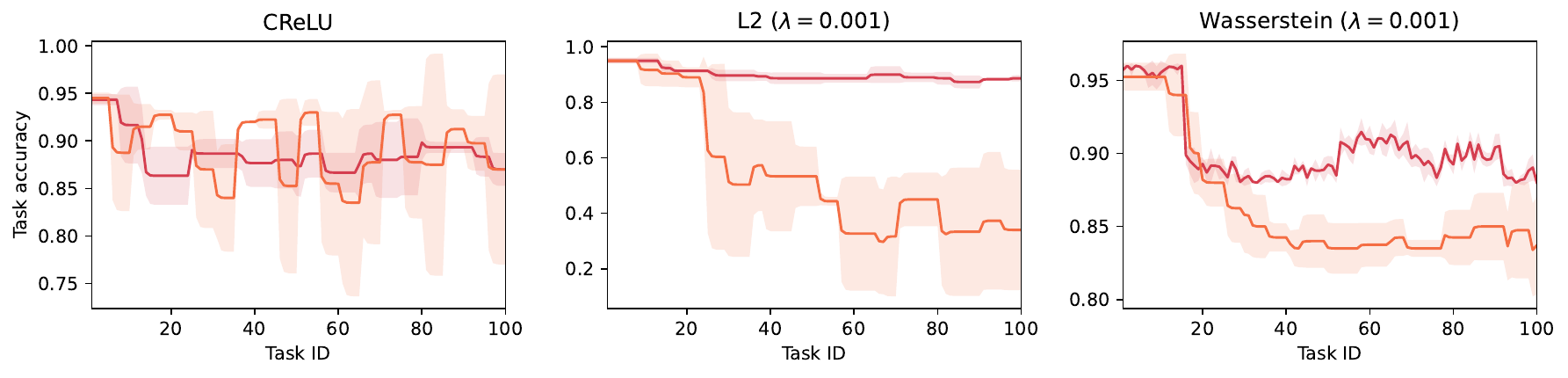}
\vspace{-1.0em}
\caption{\small \textbf{Hessian rank outcomes across methods.} The adaptive scheduler (red) helps mitigate the spectral collapse often observed in vanilla training (orange), particularly under L2 regularization where it maintains a higher effective rank throughout the task sequence.}\label{fig:hessian}
\end{figure}
\paragraph{When Hessian rank decay is predictive.}\label{app:theo-hr}
Rank decline aligns with our two-signal view when curvature mass concentrates into a low-dimensional dominant subspace but the gradient does not increasingly align with that subspace, and the per-sample gradient noise does not fall commensurately. 
Write the eigendecomposition $H_t=\sum_{i=1}^{d}\lambda_{i,t}v_{i,t}v_{i,t}^\top$ with $\lambda_{1,t}\!\ge\!\lambda_{2,t}\!\ge\dots$, let $r_t$ be an effective rank, and $P_{r_t}$ the projector onto $\{v_{1,t},\ldots,v_{r_t,t}\}$. 
Define the \emph{projected} batch-size–aware critical step-size
\[
\alpha_{g,(r)}^\star(t)\;:=\;\frac{B\,\|P_{r_t}\widehat g_t\|^2}{\sigma^{2(r)}_{t,\mathrm{ps}}}
\;\approx\;\cos^2\!\theta_t\cdot \alpha_g^\star(t)
\quad\text{(approx.\ isotropic noise),}
\]
where $\cos^2\!\theta_t=\|P_{r_t}\widehat g_t\|^2/\|\widehat g_t\|^2$. 
If $r_t\!\downarrow$ while $\cos^2\!\theta_t$ fails to rise (or falls), then $\alpha_{g,(r)}^\star(t)$ drops and a larger fraction of steps become noise-dominated, matching accuracy decay even without explicit forgetting.

In modern networks, rank decay typically coincides with spectral mass concentrating into a few outliers \parencite{Sagun2017,Papyan2018,Ghorbani2019}, which makes $\lambda_{1,t}$ more sensitive to data/parameter drift; empirically this raises $\mathrm{Vol}_{\nsharp}(t)$ (Eq.~\eqref{eq:voli}) and hence \emph{shrinks} $\alpha_{\mathrm{vol}}^\star(t)$ \emph{before} accuracy collapses. 
Studies of alignment further show gradients tend to align with a small top-eigenspace \parencite{Arora2022,Song2025}; when that space itself shrinks or drifts quickly, alignment cannot compensate, yielding a simultaneous reduction in $\alpha_{g,(r)}^\star(t)$ and $\alpha_{\mathrm{vol}}^\star(t)$. 
As $\alpha_t$ crosses either threshold persistently, LoT follows.

\subsection{Gradient Noise}\label{app:theo-grad_noise_proof}
Here we detail the derivation of $\alpha_g^\star(t)$ in Section~\ref{sec:signals}.
While Adam employs preconditioned updates, we analyze the stability of the effective step magnitude $\alpha_t$ along the chosen update direction.
Assume the loss $J$ is locally $L$-smooth along this direction.
Let the update be $w_{t+1} = w_t - \alpha_t \widehat{g}_t$, where $\widehat{g}_t$ is the stochastic update vector (e.g., the preconditioned gradient in Adam).
Let $g_t = \mathbb{E}[\widehat{g}_t]$ be the true direction and $\widehat{g}_t = g_t + \xi_t$ with noise $\xi_t$ such that $\mathbb{E}[\xi_t]=0$ and $\mathbb{E}[\|\xi_t\|^2] = \text{Var}[\widehat{g}_t]$.

A quadratic expansion of the loss yields:
\[
J(w_{t+1}) \approx J(w_t) - \alpha_t \langle \nabla J(w_t), \widehat{g}_t \rangle + \frac{1}{2}\alpha_t^2 \lambda_t \|\widehat{g}_t\|^2,
\]
where $\lambda_t$ is the curvature along the update direction ($\lambda_t \le L$). Taking expectations over the noise:
\[
\mathbb{E}[J(w_{t+1}) - J(w_t)] \le -\alpha_t \|g_t\|^2 + \frac{1}{2}\alpha_t^2 L \left( \|g_t\|^2 + \text{Var}[\widehat{g}_t] \right).
\]
For expected descent (LHS $< 0$), we require:
\[
\alpha_t \|g_t\|^2 > \frac{1}{2}\alpha_t^2 L \left( \|g_t\|^2 + \text{Var}[\widehat{g}_t] \right)
\implies \alpha_t < \frac{2}{L} \frac{\|g_t\|^2}{\|g_t\|^2 + \text{Var}[\widehat{g}_t]}.
\]
Using the per-sample variance proxy $\sigma^2_{t,\mathrm{ps}}$ where $\text{Var}[\widehat{g}_t] \approx \sigma^2_{t,\mathrm{ps}}/B$, and noting that typically $\|g_t\|^2 \ll \sigma^2_{t,\mathrm{ps}}/B$ in the noise-dominated regime, we obtain the bound proportional to the signal-to-noise ratio derived by \textcite{Schaul2013}:
\[
\alpha_t \lesssim \frac{2}{L} \frac{B \|g_t\|^2}{\sigma^2_{t,\mathrm{ps}}}.
\]

\subsection{Normalized sharpness}\label{app:theory-norm}
Consider the local quadratic model with a preconditioned step \(w_{t+1}=w_t-\eta_t D_t \widehat g_t\), where \(D_t=\mathrm{diag}\big(1/(\sqrt{\widehat v_t}+\epsilon)\big)\succ0\). The second-order term governing stability along this update is
\[
\Delta w_t^\top H_t \Delta w_t \;=\; (\eta_t \widehat g_t)^\top D_t^{1/2} H_t D_t^{1/2} (\eta_t \widehat g_t).
\]
Thus the relevant curvature scale is \(\lambda_{\max}\!\big(D_t^{1/2}H_tD_t^{1/2}\big)\). By submultiplicativity of the spectral norm,
\[
\lambda_{\max}\!\big(D_t^{1/2}H_tD_t^{1/2}\big) \;\le\; \|D_t\|_2\,\lambda_{\max}(H_t)\;=\; d_{\max,t}\,\lambda_t,
\]
where \(d_{\max,t}\) is the largest diagonal entry of \(D_t\). Replacing \(D_t\) by a scalar surrogate \(d_t I\) yields a tractable approximation
\(\lambda_{\max}\!\big(D_t^{1/2}H_tD_t^{1/2}\big)\approx d_t\,\lambda_t\).
We take \(d_t:=\alpha_{\mathrm{agg},t}/\eta_t\) with
\[
\alpha_{\mathrm{agg},t} \;=\; \frac{\eta_t}{\mathrm{RMS}(\sqrt{\widehat v_t})+\epsilon},
\]
which lies between \(\eta_t d_{\min,t}\) and \(\eta_t d_{\max,t}\). Hence
\[
\eta_t^2\,\lambda_{\max}\!\big(D_t^{1/2}H_tD_t^{1/2}\big)
\;\lesssim\; \big(\alpha_{\mathrm{agg},t}\,\lambda_t\big)\cdot \frac{d_{\max,t}}{d_t},
\]
so \(\bar\lambda_t:=\alpha_{\mathrm{agg},t}\lambda_t\) is a conservative scalar proxy up to a factor depending on the spread of \(D_t\) (bounded by its condition number). In practice this surrogate tracks the preconditioned curvature closely enough for reliable volatility detection and decision thresholds in our scheduler. A per-layer version (replacing the global RMS with layerwise RMS) is analogous and can tighten the bound further.

\subsection{Volatility}
\label{app:volatility}
\paragraph{Setup.}
With Adam, an update takes the preconditioned form
\[
w_{t+1}=w_t-\alpha_t\,\widehat g_t^{(\mathrm{eff})},\qquad 
\widehat g_t^{(\mathrm{eff})}:=D_t\,\widehat g_t,\quad
D_t=\mathrm{diag}\!\Big(\tfrac{1}{\sqrt{\widehat v_t}+\epsilon}\Big).
\]
Let \(\lambda_t=\lambda_{\max}(H_t)\). We work with the normalized (Adam-adjusted) sharpness \(\bar{\lambda}_t:=\alpha_{\mathrm{agg},t}\lambda_t\) where \(\alpha_{\mathrm{agg},t}=\eta_t/(\mathrm{RMS}(\sqrt{\widehat v_t})+\epsilon)\).
Over a short window \(W\), we treat \(\{\bar{\lambda}_s\}\) as a scalar stochastic process. We track its statistics using an exponential moving average with decay \(\nu\) (e.g., \(\nu=0.9\)):
\[
\mu_t:=\frac{(1 - \nu) \mu_{t-1} + \nu\bar{\lambda}_t}{1-\nu^t},\quad 
\sigma_t^2:=\text{EMA}_{\nu}[(\bar\lambda - \mu)^2]_t.
\]
We write the one-step quadratic model along the (preconditioned) update direction:
\[
\Delta j_s \;\lesssim\; -\alpha_t\,\|\widehat g^{(\mathrm{eff})}_s\|^2 \;+\; \tfrac{1}{2}\alpha_t^2\,\bar{\lambda}_s\,\|\widehat g^{(\mathrm{eff})}_s\|^2.
\tag{A.1}
\label{eq:quad}
\]
The first term is the \emph{deterministic} descent, the second is the \emph{curvature} penalty, whose \emph{temporal} fluctuation we will control.

\paragraph{A small-step contraction criterion via log-linearization.}
Consider the scalar dynamics of the top mode (after preconditioning): \(z_{s+1}\approx (1-\alpha_t\bar{\lambda}_s)z_s\).
For \(0\le x<1\), the inequality \(\log(1-x)\le -x - \tfrac{1}{2}x^2\) yields
\[
\mathbb{E}\big[\log|1-\alpha_t\bar{\lambda}_s|\big]
\;\le\; -\alpha_t\,\mu_t \;-\; \tfrac{1}{2}\alpha_t^2\,\mathbb{E}[\bar{\lambda}_s^2]
\;=\; -\alpha_t\,\mu_t \;-\; \tfrac{1}{2}\alpha_t^2(\mu_t^2+\sigma_t^2).
\tag{A.2}
\]
A sufficient condition for expected contraction is that the volatility (quadratic) term remain a chosen fraction \(c\in(0,1)\) of the linear term:
\[
\tfrac{1}{2}\alpha_t^2(\mu_t^2+\sigma_t^2)\ \le\ c\,\alpha_t\,\mu_t
\quad\Longrightarrow\quad
\alpha_t\ \le\ \frac{2c\,\mu_t}{\mu_t^2+\sigma_t^2}.
\tag{A.3}
\label{eq:alpha-mean-square}
\]
\emph{Noise-dominated regime.} When \(\sigma_t^2\gg \mu_t^2\), Eq.~\eqref{eq:alpha-mean-square} reduces to
\[
\alpha_t\ \lesssim\ \frac{2c\,\mu_t}{\sigma_t^2}\ =\ \frac{2c}{\mathrm{Vol}_{\bar{\lambda}}(t)}\,,
\qquad
\mathrm{Vol}_{\bar{\lambda}}(t):=\frac{\sigma_t^2}{\mu_t+\varepsilon}.
\tag{A.4}
\]
This motivates the practical curvature-driven critical step-size
\[
\ \ \alpha^{\star}_{\mathrm{vol}}(t)\ :=\ \frac{1}{\kappa\,\mathrm{Vol}_{\bar{\lambda}}(t)}\ \
\]
with \(\kappa\) absorbing the constants (the choice of \(c\), the small \(\varepsilon\), and mismatch between the scalar surrogate \(\bar{\lambda}\) and \(\lambda_{\max}(D_t^{1/2}H_tD_t^{1/2})\)).
Intuitively, Eq.~\eqref{eq:alpha-mean-square} comes from requiring the \emph{temporal} curvature fluctuation penalty to stay subdominant to the deterministic descent; in the high-volatility regime the controlling quantity is \(\sigma_t^2/\mu_t\), i.e., \(\mathrm{Vol}_{\bar{\lambda}}\).

From Eq.~\eqref{eq:quad}, a step is unstable if \(\alpha_t\bar{\lambda}_s\ge 2\).
Cantelli’s inequality gives \(\mathbb{P}(\bar{\lambda}_s-\mu_t\ge a)\le \frac{\sigma_t^2}{\sigma_t^2+a^2}\).
Setting \(a=2/\alpha_t-\mu_t\) and requiring this probability \(\le\delta\) yields a sufficient condition
\[
\alpha_t\ \le\ \frac{2}{\mu_t + \sigma_t\sqrt{(1-\delta)/\delta}}\ .
\tag{A.5}
\label{eq:cantelli}
\]
Eq.~\eqref{eq:cantelli} is tighter when \(\mu_t\) is not dwarfed by \(\sigma_t\); Eq.~\eqref{eq:alpha-mean-square} (hence \(\alpha^{\star}_{\mathrm{vol}}\propto 1/\mathrm{Vol}_{\bar{\lambda}}\)) is simpler and conservative when volatility dominates. In practice, we use \(\alpha^{\star}_{\mathrm{vol}}\) for fast control and optionally cap \(\alpha_t\) by Eq.~\eqref{eq:cantelli} for extra safety.

\subsection{Link to symmetry-induced curvature collapse}

\textcite{Ziyin2024Symmetry} demonstrate that common mirror symmetries in deep networks impose algebraic constraints on the Hessian. Under weight decay or large gradient noise, these constraints bias SGD toward symmetric, low-rank or sparse solutions, where so-called ``collapse'' phenomena emerge. In our framework, such phases correspond to decreased $\mu_t$ and/or increased $\sigma_t^2$, leading to a contraction of $\alpha^{\star}_{\mathrm{vol}}$ and signaling the onset of untrainability prior to an observable accuracy drop.

Let $H$ denote the Hessian of the loss at parameters $\theta_t$, and let $P$ be the orthogonal projector onto the subspace associated with a given symmetry (e.g., $O$-mirror symmetry). Write $d_t$ for the normalized update direction. By the Courant--Fischer theorem,
\[
\lambda_{\max}(PHP) \;=\; \max_{\substack{\|v\|=1 \\ v \in \mathrm{range}(P)}} v^\top H v \;\leq\; \lambda_{\max}(H),
\]
and analogously,
\[
\lambda_{\min}(PHP) \;\geq\; \lambda_{\min}(H).
\]

For the Ritz value $\rho_t = d_t^\top H d_t$, we thus obtain
\[
\lambda_{\min}(PHP) \;\leq\; \rho_t \;\leq\; \lambda_{\max}(PHP) \;\leq\; \lambda_{\max}(H).
\]

Define the overlap factor $\alpha_t = \|Pd_t\| / \|d_t\|$. Expanding in terms of $Pd_t$,
\[
\rho_t = (Pd_t)^\top H (Pd_t) \;\pm\; \text{cross terms}.
\]

If $d_t$ lies predominantly in the symmetry subspace (so cross terms are negligible),
\[
\rho_t \;\approx\; \alpha_t^2 \,\lambda_{\max}(PHP) \;\leq\; \alpha_t^2 \lambda_{\max}(H).
\]

As symmetry constraints ``lock in'' during training, $\alpha_t \to 1$, and the Ritz value $\rho_t$ becomes tightly bracketed by the projected and global sharpness.

Let $\sigma^2_\rho = \mathrm{Var}[\rho_t]$ and $\sigma^2_{\lambda} = \mathrm{Var}[\lambda_{\max}(H)]$. If $\alpha_t$ varies slowly and remains close to one, then
\[
\sigma^2_\rho \;\approx\; \sigma^2_{\lambda}.
\]

Hence, fluctuations in $\lambda_{\max}(H)$ provide a practical proxy for the volatility of the stability-critical curvature.

Finally, \textcite{Ziyin2024Symmetry} derive a Lyapunov-type stability criterion in symmetry-restricted directions, expressed in terms of the mean and variance of curvature:
\[
\text{Collapse if:} \quad \eta \,\bar{\xi} \;\gtrsim\; \frac{-2\,\mathbb{E}[\xi+\gamma]}{\mathbb{E}[(\xi+\gamma)^2]}.
\]

Substituting $\xi \mapsto \rho_t$ and applying the above bounds yields a conservative but practical collapse rule:
\[
\eta_t \cdot \lambda_{\max}(H) \;\lesssim\; \text{margin}(\text{mean}, \text{var}).
\]

\section{Implementation details}\label{app:impl}
\subsection{Metrics}
\label{app:impl-metrics}

\paragraph{Effective step-size (global \& per-layer).}
We report the \emph{effective step-size} $\alpha_t$ actually applied by the optimizer.
For Adam/ClampedAdam we average, over parameters, the elementwise multiplier
\[
  \frac{\eta_t}{(1-\beta_1^t)\,(\sqrt{\hat v_t}+\varepsilon)}\,.
\]
We also compute a \emph{per-layer} effective step-size $\alpha_{t,\ell}$ by aggregating within each top-level module (e.g., \texttt{fc1}, \texttt{fc2}). These $\alpha_t$/$\alpha_{t,\ell}$ are compared against the safety thresholds below.

\paragraph{Sharpness and normalized sharpness.}
We estimate the top Hessian eigenvalue $\lambda_t$ via power iteration with Hessian vector products ($k{=}1$, $\sim$100 steps).
For Adam we form a \emph{normalized sharpness} scalar
\[
  \bar\lambda_t \;=\; \alpha_{\mathrm{agg},t}\,\lambda_t,
  \qquad
  \alpha_{\mathrm{agg},t} \;=\; \frac{\eta_t}{\mathrm{RMS}(\sqrt{\hat v_t}) + \varepsilon},
\]
which tracks the preconditioned curvature along the update; for SGD we set $\bar\lambda_t=\lambda_t$.
For each layer $\ell$ we maintain windowed statistics of $\bar\lambda_{t,\ell}$ using an EMA plus a finite queue (length $30$):
the running mean $\mu_{t,\ell}$ and the instantaneous squared deviation $(\bar\lambda_{t,\ell}-\mu_{t,\ell})^2$, which we use for the volatility threshold in Eq.~\ref{eq:voli}.

\paragraph{Batch-size--aware gradient noise.}
At log intervals we compute the exact within-minibatch per-sample gradient variance
\[
  \hat\sigma^2_{\mathrm{mb}}(t) \;=\; \frac{1}{B}\sum_{i=1}^{B} \bigl\| g_i - \bar g \bigr\|^2,
\]
by looping over per-sample losses. 

\paragraph{Hessian rank.}
Following \textcite{Lewandowski2023}, we approximate the local curvature complexity using the effective rank of the Empirical Fisher Information Matrix (EFIM).
Given a subset of $M$ samples (here $M=100$), we collect per-sample gradients into a matrix $G \in \mathbb{R}^{M \times P}$, where $P$ is the number of parameters.
To avoid the computational cost of the full $P \times P$ matrix, we compute the spectrum of the Gram matrix $K = GG^\top \in \mathbb{R}^{M \times M}$.

\subsection{Experiments}\label{app:impl-expl}
In Fig.~\ref{fig:predict_controller}(a), we compute per-layer predictions as follows: for each layer $\ell$, we evaluate the critical step-size $\tilde{\alpha}_t^{\star(\ell)}(t)$ and calculate a predicted LoT as:
\[
\widehat{\rho}\;=\;\frac{1}{T}\sum_{t=1}^{T}\mathbf{1}\!\left\{\exists\,\ell:\ \alpha_t^{(\ell)}>\widetilde{\alpha}_t^{\star(\ell)}(t)\right\}.
\]
Since $\widehat\rho$ only captures the relative training behavior, we scale it to fit the range of the actual trainability curve.

\subsection{Algorithm}
Refer to \ref{app:setup-hyper} for the model and hyperparameter specifications.
\begin{algorithm2e}
\caption{\textbf{Per-layer sharpness-aware LR scheduler (ours), invoked every $K$ steps}}
\label{alg:scheduler}
\DontPrintSemicolon
\KwIn{layer params \(w^{(\ell)}\), base LR \(\eta^{(\ell)}\), EMA state \(\mathcal{E}^{(\ell)}\), decision-window \(W\), safety factor \(\gamma\), cool rate \(c=1-\varepsilon\), warm rate \(u=1+\varepsilon\); \(\varepsilon\approx 10^{-2}\), interval \(K\)}

\For{\(t \in \{K,2K,3K,\dots\}\) until \(t \ge T\)}{
    Estimate per-layer sharpness \(\lambda_t^{(\ell)}\) (top Hessian eigenvalue), calculate normalized sharpness \(\bar\lambda_t^{(\ell)}\), and minibatch grad variance \(\sigma^{2(\ell)}_{t,\mathrm{mb}}\)\;
  Update EMA state \(\mathcal{E}^{(\ell)}\) with \((\lambda_t^{(\ell)}, \alpha_t^{(\ell)})\)\;
  Compute \(\tilde\sigma_{t}^{2(\ell)}\), \(\tilde\alpha_{t}^{\star(\ell)}\), and safe bound \(\tilde\alpha_{t}^{\mathrm{safe}(\ell)}={(1-\epsilon)}\tilde\alpha_{t}^{\star(\ell)}\) \tcp*{via Eq.~\eqref{eq:alpha-tilde-layer}}
  Form effective step-size \(\alpha_t^{(\ell)}\) from optimizer state 
    
  \eIf{\(\alpha_t^{(\ell)} >  \tilde\alpha_{t}^{\mathrm{safe}(\ell)}\) \textbf{and} \(\alpha_t^{(\ell)} > 0.12\)}{
    \(\eta^{(\ell)} \gets c \cdot \eta^{(\ell)}\) \tcp*{cool overly aggressive layer}
  }{
    \If{\(t < 0.3T\) \textbf{and} \(\alpha_t^{(\ell)} \ll \tilde\alpha_{t}^{\mathrm{safe}(\ell)}\)}{
      \(\eta^{(\ell)} \gets u \cdot \eta^{(\ell)}\) \tcp*{warm timid layer early on}
    }
  }
}
\end{algorithm2e}

\newpage 
\section{Full experimental setup}\label{app:setup}
Experiments use a random-label MNIST task sequence of 40 tasks with 250 epochs per task; scheduler decisions are taken every \(K\) optimizer steps with small warming/cooling updates, and power-iteration budgets are kept short to bound overhead . With longer tasks, our controller maintains trainability. 
\subsection{Dataset}
All figures (unless stated otherwise) use a random-label MNIST stream. Starting from the MNIST training split, we randomly subsample $21{,}000$ images once and normalize inputs using the dataset mean and standard deviation returned by our loader. At the beginning of each task we resample targets with full dataset randomization, so input statistics remain MNIST-like while class semantics are destroyed. For the controller experiments we run a stream of $40$ tasks with $250$ epochs per task; for the ablations we keep everything else fixed and vary the per-task budget across $100$, $250$, and $500$ epochs. All runs are averaged over $3$ seeds.

\subsection{Models}
Unless specified, the backbone is a two-layer MLP with width of $256$.

\subsection{Hyperparameters}\label{app:setup-hyper}
\begin{table}[H]
\centering
\small
\begin{tabular}{@{}ll@{}}
\toprule
\textbf{General Hyperparameters} & \textbf{Value} \\
\midrule
Label randomization & Fraction $\texttt{ns}=1.0$ (full dataset re-label each task) \\
Batch size & $256$ \\
Model & 2-layer MLP (hidden width $256$) \\
Initialization & Kaiming (default) \\
Optimizer & Adam \\
$\beta_1$ & $0.9$ \\ 
$\beta_2$ & $0.999$ \\
Base learning rate & $\eta=10^{-3}$ ($10^{-2},~10^{-4},~10^{-5}$ when specified)\\
L2 weight decay ($\lambda_{\mathrm{L2}}$) & $10^{-3}$ \\
Wasserstein reg. ($\lambda_{\mathrm{Wass}}$) & $10^{-3}$ \\
Leaky-ReLU factor ($\rho$) & $0.7,~0.3$ \\
Safety factory $\gamma$ & $0.8$ \\
Cool rate $c$ & $0.99$  \\
Warm rate $u$ & $1.01$  \\
Logging interval & $40$ \\
Controlling interval ($K$) & $40$ \\
\bottomrule
\end{tabular}
\caption{Hyperparameters used across experiments. Unless noted, the controller experiments use these settings; ablations vary only the indicated field(s).}
\label{tab:hyper}
\end{table}

\begin{table}[H]
\centering
\small
\label{tab:hyper_fig1}
\begin{tabular}{@{}ll@{}}
\toprule
\textbf{Figure 1(a) Hyperparameters} & \textbf{Value} \\
\midrule
Learning rate & $10^{-4}$ \\
L2 weight decay ($\lambda_{\mathrm{L2}}$) & $10^{-3}$ \\
Wasserstein reg. ($\lambda_{\mathrm{Wass}}$) & $10^{-3}$ \\
\bottomrule
\\
\toprule
\textbf{Figure 1(b) Hyperparameters} & \textbf{Value} \\
\midrule
Learning rate & $10^{-3}$ \\
L2 weight decay ($\lambda_{\mathrm{L2}}$) & $10^{-5}$ \\
Leaky-ReLU factor ($\rho$) & $0.3$ \\
\bottomrule
\\
\toprule
\textbf{Figure 1(c), (d) Hyperparameters} & \textbf{Value} \\
\midrule
Learning rate & $10^{-3}$ \\
Leaky-ReLU factor ($\rho$) & $0.3,~0.7$ \\
\bottomrule
\\
\toprule
\textbf{Figure 1(e) Hyperparameters} & \textbf{Value} \\
\midrule
Learning rate & $10^{-5}$ \\
L2 weight decay ($\lambda_{\mathrm{L2}}$) & $10^{-3}$ \\
Wasserstein reg. ($\lambda_{\mathrm{Wass}}$) & $10^{-3}$ \\
\bottomrule
\\
\toprule
\textbf{Figure 1(f) Hyperparameters} & \textbf{Value} \\
\midrule
Learning rate & $10^{-3}$ \\
Leaky-ReLU factor ($\rho$) & $0.01$ \\
\bottomrule
\end{tabular}
\caption{Method-based hyperparameters used across experiments in Figure 1. Other hyperparameters are the same as mentioned in Table \ref{tab:hyper}.}
\end{table}

\end{document}